# RANSAC based three points algorithm for ellipse fitting of spherical object's projection

Shenghui Xu

Beihang University

xshhhm@sa.buaa.edu.cn

**Abstract**

As the spherical object can be seen everywhere, we should extract the ellipse image accurately and fit it by implicit algebraic curve in order to finish the 3D reconstruction. In this paper, we propose a new ellipse fitting algorithm which only needs three points to fit the projection of spherical object and is different from the traditional algorithms that need at least five point. The fitting procedure is just similar as the estimation of Fundamental Matrix estimation by seven points, and the RANSAC algorithm has also been used to exclude the interference of noise and scattered points.

# 1 Introduction

The ellipse fitting problem that is a part of curve fitting problem can be described like this: Fit the points in the plane coordinate using the ellipse equation. There are various algorithms to solve the ellipse fitting problem: SVD decomposition [14,15] is the generally solution; A Andrew Fitzgibbon et al. [1] proposed a direct least square algorithm; Paul L. Rosin [6] utilized the LMedS method to estimate the ellipse parameters robustly; Daniel Keren and Craig Gotsman [5] added a circle constraints to fit curves and surfaces; Sung Joon Ahn et al. [2] chose the orthogonal distances to fit the ellipse; Tsuyoshi and Ryo-ich [7] used the genetic algorithm; F. Mai et al. [8] made use of RANSAC algorithm to extract ellipse; Markus Vincze [9] used a probabilistic method (RANSAC-like) to track ellipse.

As there are some special properties of the projection of spherical object [3, 4, 13]: the freedom of ellipse is three. Only three points are needed to estimate the ellipse parameters. We will prove the ellipse fitting based on RANSAC is similar as the estimation of Fundamental Matrix [11, 14].

# 2 Ellipse fitting problem

## 2.1 Least square fitting

Many curve fitting problems can be solved by least square. We use the polynomial equation to fit the N points just like:

$$f(x,y) = \sum_{0 \leq i+j \leq n} a_{i,j} x^i y^j = a_{n,0} x^n + a_{n-1,1} x^{n-1} y + a_{n-2,2} x^{n-2} y^2 + \cdots + a_{i,j} x^i y^j + \ldots + a_{0,1} y + a_{0,0} = 0 \tag{1}$$

Given the coordinate value $(x_i, y_i)$ of each point, the equation (1) can be converted to the following matrix form:

$$f(x_i, y_i) = a_{n,0}x_i^n + a_{n-1,1}x_i^{n-1}y_i + \cdots + a_{i,j}x_i^i y_i^j + \ldots + a_{0,1}y_i + a_{0,0} = \begin{bmatrix} x_i^n & x_i^{n-1}y_i & \cdots & x_i^i y_i^j & \cdots & y_i & 1 \end{bmatrix} \begin{bmatrix} a_{n,0} \\ a_{n-1,1} \\ \vdots \\ a_{i,j} \\ \vdots \\ a_{0,1} \\ a_{0,0} \end{bmatrix} = 0 \quad (2)$$

Every point $(x_i, y_i)$ satisfies the equation (2), so the N points meet the form:

$$\begin{bmatrix} x_1^n & x_1^{n-1}y_1 & \cdots & x_1^i y_1^j & \cdots & y_1 & 1 \\ x_2^n & x_2^{n-1}y_2 & \cdots & x_2^i y_2^j & \cdots & y_2 & 1 \\ \vdots & \vdots & \vdots & \vdots & \vdots & \vdots & \vdots \\ x_N^n & x_N^{n-1}y_N & \cdots & x_N^i y_N^j & \cdots & y_N & 1 \end{bmatrix} \begin{bmatrix} a_{n,0} \\ a_{n-1,1} \\ \vdots \\ a_{i,j} \\ \vdots \\ a_{0,1} \\ a_{0,0} \end{bmatrix} = Ap = 0, \quad (3)$$

in which $A = \begin{bmatrix} x_1^n & x_1^{n-1}y_1 & \cdots & x_1^i y_1^j & \cdots & y_1 & 1 \\ x_2^n & x_2^{n-1}y_2 & \cdots & x_2^i y_2^j & \cdots & y_2 & 1 \\ \vdots & \vdots & \vdots & \vdots & \vdots & \vdots & \vdots \\ x_N^n & x_N^{n-1}y_N & \cdots & x_N^i y_N^j & \cdots & y_N & 1 \end{bmatrix}$, $p = \begin{bmatrix} a_{n,0} \\ a_{n-1,1} \\ \vdots \\ a_{i,j} \\ \vdots \\ a_{0,1} \\ a_{0,0} \end{bmatrix}$.

The fitting of a general conic may be approached by minimizing the sum of squared algebraic distances

$$D = \|Ap\| = (Ap)^T(Ap) = p^T(A^T A)p \\ s.t. \|p\| = p^T p = 1 \quad (4)$$

from the curve to the N data points $(x_i, y_i)$.

The purpose of curves fitting is to estimate the optimal value of vector $p$. Generally speaking, solving the eigen value of matrix $A^T A$, the eigenvector corresponds to the minimum eigenvalue is the estimation value of $p$ [14].

### 2.2 Ellipse fitting

The ellipse fitting is the special situation of equation (2). We can represent the ellipse curve using function

$$f(x, y) = \sum_{0 \le i+j \le 2} a_{i,j}x^i y^j = a_{2,0}x^2 + a_{1,1}xy + a_{0,2}y^2 + a_{1,0}x + a_{0,1}y + a_{0,0} = 0 \quad (5)$$

The equation (5) can be denoted as:

$$f(x,y) = ax^2 + 2bxy + 2dx + cy^2 + 2ey + f = X^T CX = [x^2 \quad 2xy \quad y^2 \quad 2x \quad 2y \quad 1]\begin{bmatrix} a \\ b \\ c \\ d \\ e \\ f \end{bmatrix} = 0 \qquad (6)$$

$$X = \begin{bmatrix} x \\ y \\ 1 \end{bmatrix}, P = \begin{bmatrix} a & b & d \\ b & c & e \\ d & e & f \end{bmatrix}$$

Where $X$ represents the homogeneous coordinate of points, and $P$ represents the coefficient matrix of ellipse.

If there are N points waiting for fitting, we can attain the similar form like equation (3) and (4):

$$D = \|Ap\|^2 = (Ap)^T (Ap) = p^T A^T A p$$

$$s.t \|p\| = 1, p = \begin{bmatrix} a \\ b \\ c \\ d \\ e \\ f \end{bmatrix} \qquad (7)$$

where $p$ represents the ellipse coefficients.

We can estimate the parameters $p$ by SVD or Eigen value decomposition [13]. But this estimation algorithm didn't take use of the ellipse constraint condition

$$\Delta = \det(\begin{bmatrix} a & b \\ b & c \end{bmatrix}) > 0 \qquad (8)$$

When the noise exists, the fitting result that may be hyperbola or parabola, just not ellipse is unstable. To solve this problem, Andrew Fitzgibbon et al. [1] proposed a new direct least square fitting of ellipses algorithm which can force the fitting result be ellipse.

## 2.3 Ellipse fitting for the projection of spherical object

The freedom of ellipse is five, so only five points are needed for the ellipse fitting. But for the special situation of spherical object, the projection image is an ellipse which satisfies two constraints [3, 4, 13] as follow:

$$\begin{aligned} S_1 &= d(bd - ae) - e(be - cd) = 0 \\ S_2 &= b(ae - bd) f_e^2 - e(de - bf)(l^2 - 1) = 0 \end{aligned} \qquad (9)$$

where $l = 0$ in the pinhole camera model, $f_e$ is the focal length.

The first constraint $S_1$ represent that the main axis of ellipse walk through the centre of image.

The second constraint denotes the relationship between the intrinsic parameter matrix $K$ and coefficients matrix $P$.

So the freedom of spherical object's projection image is three, we can just choose three points to estimate the ellipse parameters.

# 3 RANSAC algorithm

The fitting result of ellipse will be not stable as the existence of noise and scattered points. The RANSAC algorithm [10, 12] is a good choice to apply in our fitting procedure. In this section, we will prove that the fitting procedure of ellipse is just similar as the estimation of Fundamental Matrix estimation by seven points using RANSAC algorithm.

## 3.1 Introduction to RANSAC algorithm

The RANSAC algorithm is an abbreviation for "RANdom Sample Consensus" which was first published by Fischler and Bolles in 1981 and is an iterative method to estimate parameters of a mathematical model from a set of observed data which contains outliers.

Generally fitting of ellipse by RANSAC only need five sample points, and the process is as follows:

---

Step 1    Choose five points randomly from the observed points to fit the ellipse model.

Step 2    Calculate the distances from the points to the fitting model, then discriminate the distances whether in the threshold or not, if less than the threshold the points will be labeled as inliers. Define the set of liners as consistent set.

Step 3    Compute the number of inliers set. If the number is less than the defined threshold, go to step 1, if greater, end the loop.

Step 4    Choose the consistence set with the largest number of inliers.

Step 5    If the iteration number exceeds the maximum number, stop the loop.

Step 6    Re-estimate the model using the inliers with the largest number.

---

Because the freedom of projection ellipse is three, only three sample points are needed to fit the ellipse model.

## 3.2 Application of RANSAC in ellipse fitting

For the ellipse fitting, only three sample points are required and they satisfy the equation

$$\begin{bmatrix} x_1^2 & 2x_1y_1 & y_1^2 & 2x_1 & 2y_1 & 1 \\ x_2^2 & 2x_2y_2 & y_2^2 & 2x_2 & 2y_2 & 1 \\ x_3^2 & 2x_3y_3 & y_3^2 & 2x_3 & 2y_3 & 1 \end{bmatrix} \begin{bmatrix} a \\ b \\ c \\ d \\ e \\ f \end{bmatrix} = Ap = 0 \qquad (10)$$

in which the rank of $A$ is 3, through SVD decomposition we can get

$$A = U_{3 \times 3} diag(\sigma_1, \sigma_2, \sigma_3, 0, 0, 0) V_{6 \times 6} \qquad (11)$$

then $A(V_4 V_5 V_6) = 0_{3 \times 3}$ \qquad (12)

Define a mapping between $p$ and $P$:

$$p\text{-> } P\!: \quad p = \begin{bmatrix} a \\ b \\ c \\ d \\ e \\ f \end{bmatrix} \text{->} \quad P = \begin{bmatrix} a & b & d \\ b & c & e \\ d & e & f \end{bmatrix} \tag{13}$$

where $p$ represents the coefficient vector, $P$ denotes the coefficient matrix.

Let $p_1 = V_4, p_2 = V_5, p_3 = V_6$, and $p_1\text{->}P_1, p_2\text{->}P_2, p_3\text{->}P_3$, then we obtain a cubic polynomial in

$$P = P_1 + \alpha P_2 + \beta P_3 \tag{14}$$

Convert the equation (9) as follows:

$$\det(S_1\_new) = 0$$
$$\det(S_2\_new) = 0$$
$$S_1\_new = \begin{bmatrix} a & b & d \\ b & c & e \\ e & -d & 0 \end{bmatrix} \tag{15}$$
$$S_2\_new = \begin{bmatrix} a & -e(l^2-1) & d \\ b & 0 & e \\ d & -bf_e^2 & f \end{bmatrix}$$

Thus, substitute equation (14) to equation (15), we can get two cubic equations, and the solution number is nine. Excluding the imaginary root, then the rest real roots $\alpha, \beta$ are what we need.

After estimating the ellipse model by three sample points, divide the rest points into inliers and outliers. The whole process is as follows:

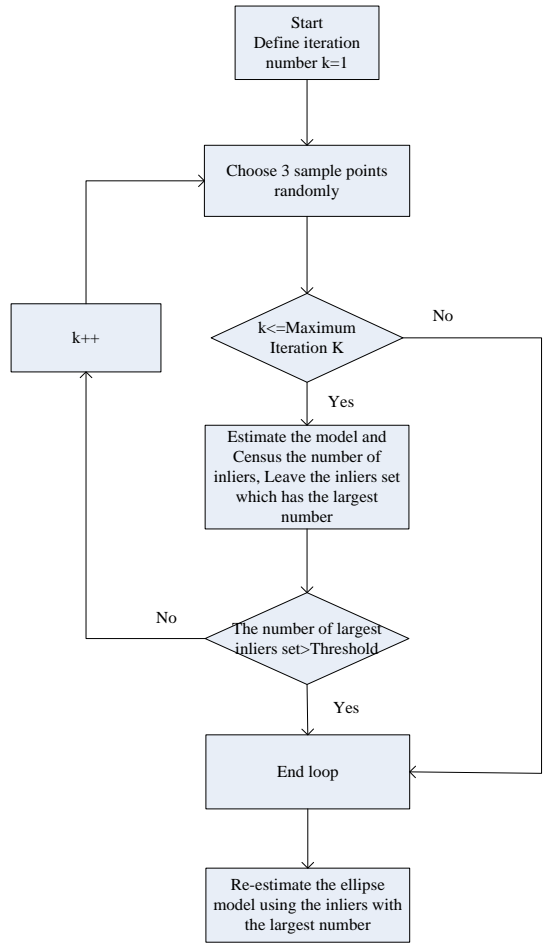

**Figure 1: the diagram of proposed algorithm**

We can find out that above ellipse fitting process is so much similar like the computation of fundamental matrix by RANSAC algorithm. But there are still several different places. One of the differences is the fitting data shouldn't be normalized, which can be proved. The proof process is omitted due to the space limitations.

# 4 Experiments

In this section, we will perform the simulation experiments in two different situations:
1   Fitting the ellipse under the influence of noise and scattered points;
2   Fitting the ellipse without scattered points, but only with noise;
   The algorithms we will implement are as follow:
   - SVD decomposition algorithm: which only takes use of the constraint 2-norm of parameters vector  $p$  is one [14, 15].
   - Direct least square fitting algorithm: which was proposed by Andrew Fitzgibbon et al. [1], and only takes use of the constraint that the 2-norm of parameters vector is one
   - RANSAC based three points algorithm: which is proposed in this paper

## 4.1 Fitting the ellipse under the influence of noise and scattered points

The ellipse fitting result is sensitive to the interference of scattered points and noise, if estimate the coefficients of ellipse with all the points, the result will be awful. We can take advantage of the

RANSAC based three points algorithm to exclude the obstacle, and the detail algorithm process refers to figure 1.

In figure 2, after several times of iteration, the scattered points have been cleared, and the rest inliers can fit the ellipse robustly.

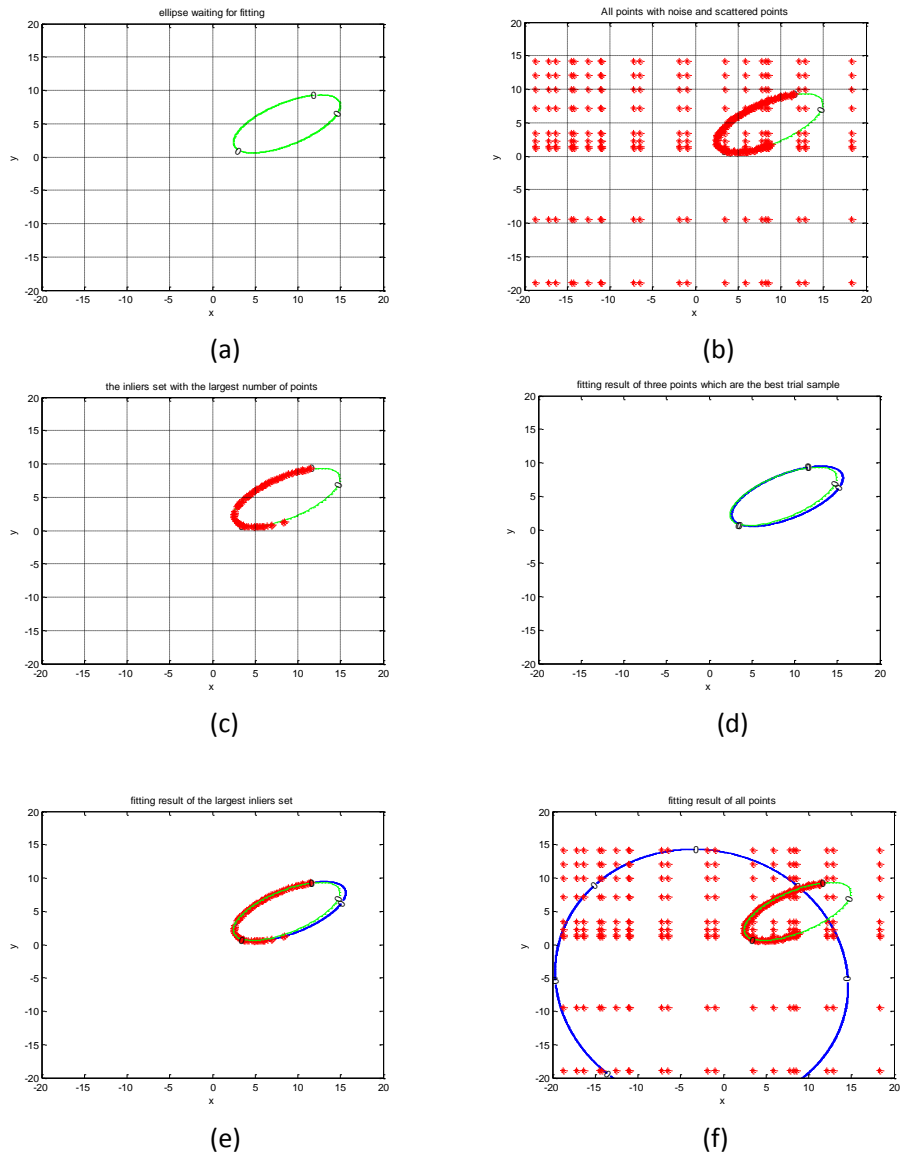

**Figure 2: Green curves are the ideal ellipse, blue are the fitting result.**

(a) Ideal ellipse;
(b) All the points with interference of scattered points and noise, the ratio of hypothetical inliers is 49%(the sum of all points is 390), and the standard deviation of noise is 0.1;
(c) Inliers set with the largest number of points, exclude the interference of scattered points;
(d) Fitting result of three sample points which are the best trial sample;
(e) Re-estimate result of the largest inliers set;
(f) Fitting result of all points.

## 4.2 Fitting the ellipse without scattered points, but only with noise

The projection of spherical object is ellipse. We want to fit the ellipse accurately. Unfortunately, there are not enough points using for fitting the ellipse because of occlusion or the interference of noise. At the same time, the fitting result is unstable. We can solve this problem by adding two constraints of equation (9), and the fitting result is robust as follows.

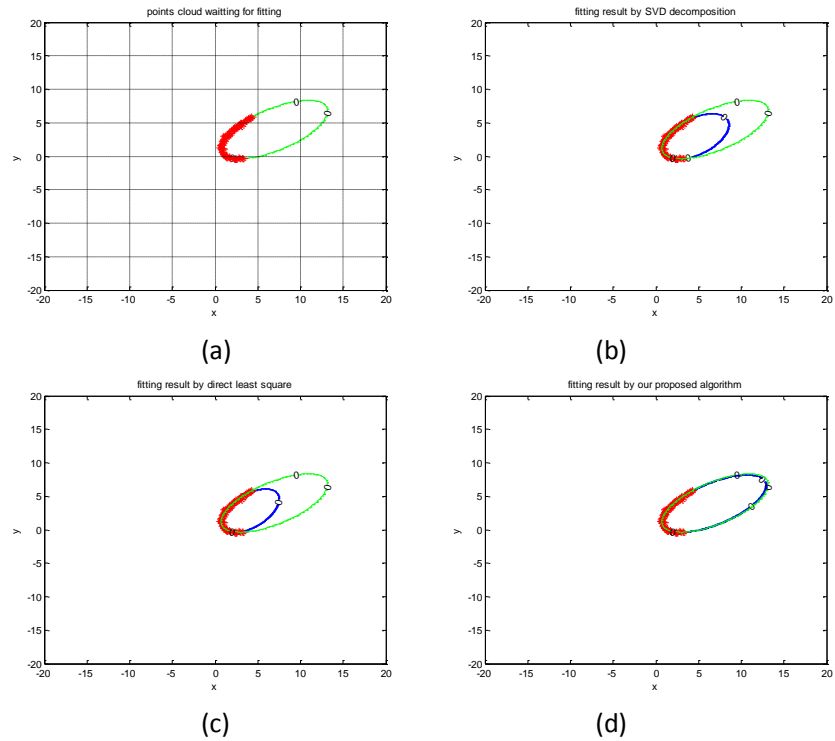

**Figure 3: Green curves are the ideal ellipse, blue are the fitting result;**

(a) All the points with the standard deviation of noise is 0.1;
(b) Fitting result by SVD decomposition;
(c) Fitting result by direct least square;
(d) Fitting result by proposed algorithm which utilize the two constraints of equation (9).

# 5 Conclusion and Discussion

As the spherical object can be seen everywhere, we should extract the ellipse image accurately and fit it by implicit algebraic curve in order to finish the 3D reconstruction. In this paper, we propose a new RANSAC based ellipse fitting algorithm which only needs three points to fit the projection of spherical object. The fitting result is accurate and robust.